\documentclass[11pt]{article}
\usepackage{coling2020}
\usepackage{times}
\usepackage{url}
\usepackage{latexsym}

\usepackage{graphicx}
\usepackage{multirow}
\usepackage{mathtools}

\colingfinalcopy % Uncomment this line for the final submission

\title{Recurrent multiple shared layers in Depth for Neural Machine Translation}

\author{GuoLiang Li \\
  KingSoft AI Lab  \\

  {\tt lyz038015@163.com} \\\And
  Yiyang Li \\
  University of Groningen \\

  {\tt adela@cau.edu.cn} \\}

\date{}

\begin{document}
\maketitle
\begin{abstract}
Learning deeper models is usually a simple and effective approach to improve model performance, but deeper models have larger model parameters and are more difficult to train. To get a deeper model, simply stacking more layers of the model seems to work well, but previous works have claimed that it can't benefit the model. We propose to train a deeper model with recurrent mechanism, which loops the encoder and decoder blocks of Transformer in the depth direction. To address the increasing of model parameters, we choose to share parameters in different recursive moments. We conduct our experiments on WMT16 English-to-German and WMT14 English-to-France translation tasks, our model outperforms the shallow Transformer-Base/Big baseline by $0.35,1.45$ BLEU points, which is $27.49\%$ of Transformer-Big model parameters. Compared to the deep Transformer(20-layer encoder, 6-layer decoder), our model has similar model performance and infer speed, but our model parameters are $54.72\%$ of the former.
\end{abstract}

\section{Introduction}
\label{intro}

%
% The following footnote without marker is needed for the camera-ready
% version of the paper.
% Comment out the instructions (first text) and uncomment the 8 lines
% under "final paper" for your variant of English.
%
\blfootnote{
    %
    % for review submission
    %
    \hspace{-0.65cm}  % space normally used by the marker
    Place licence statement here for the camera-ready version.
    %
    % % final paper: en-uk version
    %
    % \hspace{-0.65cm}  % space normally used by the marker
    % This work is licensed under a Creative Commons
    % Attribution 4.0 International Licence.
    % Licence details:
    % \url{http://creativecommons.org/licenses/by/4.0/}.
    %
    % % final paper: en-us version
    %
    % \hspace{-0.65cm}  % space normally used by the marker
    % This work is licensed under a Creative Commons
    % Attribution 4.0 International License.
    % License details:
    % \url{http://creativecommons.org/licenses/by/4.0/}.
}
Neural Machine Translation(NMT), which is a multiple layers end-to-end structure, has achieved state-of-the-art performances in large-scale translation tasks. The neural network consists of encoder and decoder, the encoder encodes the input sentence into a sequence of distributed representations, based on which the decoder generates the translation with an attention model  \cite{bahdanau2014neural,luong2015effective}. More recently, the system based on self-attention \cite{vaswani2017attention} is rapidly becoming the standard component in NMT, it demonstrates both superior performance and training speed compared to previous architectures using recurrent neural network \cite{wu2016google}.

For vanilla Transformer, training deep networks has always been a challenging problem, mainly due to the difficulties in optimization for deep architecture. In the work of \cite{He2015Delving,simonyan2014very}, they propose to train deeper model via advanced initialization schemes and multi-stage training strategy in computer vision tasks. For neural machine translation task, learning a deeper model is helpful to improve model performace, some promising attempts have proven to be of profound value. \cite{BapnaTraining} trained a 16-layer Transformer encoder by using an enhanced attention model, \cite{wu2019depth} propose an effective two-stage approach with three specially designed components to build deeper NMT model, \cite{zhang2019improving} propose depth-scaled initialization to decrease parameter variance at the initialization stage and reduce output variance of residual connections. In this work, we continue the line of research and go towards a more efficient approach to train deeper model. we focus on the encoder in our study, because encoder have a greater impact on performace than decoder and it requires less computational cost. At the same time, we use the parameters-sharing strategy to reduce the size of model. Our contributions are threefold:

\begin{itemize}
	\item Inspired by the approaches of training deeper model for neural machine translation in numerical analysis \cite{wu2019depth,wang2019learning,zhang2019improving,BapnaTraining}, we propose an approach based on recurrent mechanism to build and training a deeper model. We loop the encoder and decoder blocks of Transformer in the depth direction, every block consists of $n$ layers. Stacking more blocks simply and build a closed parameters chain are the main strategies in our experiments, which will be described in detail in Section~\ref{sec:approach}. To reduce the model parameters, we choose to share parameters in different recursive moments as well.
	\item The proposed approach is efficient and easy to implement, and it is superior to the baseline models with fewer model parameters.
	\item Our work is among the few studies \cite{wu2019depth,wang2019learning,zhang2019improving} which prove that the idea of training deeper model can have promising applications on natural language processing tasks.
\end{itemize}
We evaluate our approach on WMT16 English-to-German(En-De) and WMT14 English-to-France(En-Fr) translation tasks, and we employ the Transformer-Base/Big \cite{vaswani2017attention} and deep Transformer\cite{wang2019learning} as the strong baseline models. Experimental results show that our approach contributes to train deeper model for different translation tasks. For En-De translation task, it outperforms the Transformer-Base/Big model by $0.34,1.45$ BLEU points with $23.27\%$ model parameters. In addtion to, our model has similar model performance and infer speed, but its model parameters are $54.72\%$ of the deep Transformer(20-layer encoder, 6-layer decoder). For En-Fr translation task, it outperforms the Transformer-Base/Big model by $0.34,1.45$ BLEU points with $23.27\%$ model parameters as well.

\section{Related Work}
{\bf deeper model} Training deeper model with multiple stacked layers is challenging. Concerning nature language processing tasks, \cite{wang2019learning} have shown that the deep Transformer model can be easily optimized by proper use of layer normalization, they use pre-LN strategy to train a deeper model with 20-layer encoder and 6-layer decoder, which outperforms the Transformer-Big model and has faster training/infer speed. \cite{zhang2019improving} propose depth-scaled initialization to train a deeper model, which decreases parameter variance at the initialization stage and reduces output variance of residual connections. In the work of \cite{BapnaTraining}, they point out that vanilla Transformer is hard to train if the depth of the model is beyond12, they successfully train a 16-layer Transformer by attending the combination of all encoder layers to the decoder.

{\bf parameters-sharing} Although existing works have achieved remarkable progress in training more deeper models, while the number of model parameters increases as the model deepens. To reduce the number of model parameters, parameter-sharing is an efficient and simple strategy. In the work of \cite{LanALBERT}, they reduce the number of model parameters and increase the training speed of BERT\cite{DevlinBERT} with cross-layer parameter-sharing, their best model outperforms the BERT in many downstream tasks, and it has fewer parameters compared to BERT-large. In the work of \cite{he2018layer}, they design a layer-wise attention mechanism and further share the parameters of each layer between the encoder and decoder to regularize and coordinate the learning, parameter-sharing strategy is contributing to training a deeper model with the number of model parameters unchanged. In our work, we propose to integrate the parameter-sharing strategy to the recurrent mechanism, which can train a deeper model with fewer model parameters.

\section{Background}
\subsection{Neural Machine Translation}
The encoder-decoder framework has noticeable effect on neural machine translation tasks \cite{sutskever2014sequence,wu2016google}. Generally, multi-layer encoder and decoder are employed to perform the translation task through a series of nonlinear transformations. The encoder reads the source sentence denoted by $X=(x_{1}, x_{2}, ..., x_{M})$ and maps it to a continuous representation $Z=(z_{1}, z_{2}, ..., z_{M})$. Given $Z$, the decoder generates the target translation $Y=(y_{1}, y_{2}, ..., y_{N})$ conditioned on the sequence of tokens previously generated. The encoder-decoder framework model learns its parameter $\theta$ by maximizing the log-likelihood $p(Y|X;\theta)$, which is usually decomposed into the product of the conditional probability of each target word.
\begin{equation}
p(Y|X;\theta) = \prod_{t=1}^{N}{p(y_{t}|y_{1},...,y_{t-1},X;\theta)}
\end{equation}
where $N$ is the length of target sentence. Both the encoder and decoder can be implemented by different structure of neural models, such as RNN \cite{cho2014learning}, CNN \cite{gehring2017convolutional} and self-attention \cite{vaswani2017attention}.
\subsection{Self-attention based network}
Self-attention networks \cite{shaw2018self,so2019evolved} have attracted increasing attention due to their flexibility in parallel computation and dependency modeling. Self-attention networks calculate attention weights between each pair of tokens in a single sequence, and then a position-wise feed-forward network to increase the non-linearity. The self-attention is formulated as:
\begin{equation}
Attention(Q,K,V) = softmax(\frac{QK^{T}}{\sqrt{d_{model}}})V
\end{equation}
where $Q,K,V$ denotes the query, key and value vectors, $d_{model}$ is the dimension of hidden representations. The position-wise feed-forward network consists of a two-layer linear transformation with ReLU activation in between:
\begin{equation}
FFN(x) = max(0, xW_{1} + b_{1})W_{2} + b_{2}
\end{equation}
Instead of performing a single attention function, it is beneficial to capture different context features with multiple individual attention functions, namely multi-head attention. Specifically, multi-head attention model first transforms $Q,K$ and $V$ into $H$ subspaces with different linear projections:
\begin{equation}
Q_{h}, K_{h}, V_{h} = QW_{h}^{Q}, KW_{h}^{K}, VW_{h}^{V}
\end{equation}
where $Q_{h}, K_{h}, V_{h}$ are respectively the query, key and value representation of the $h$ head, $W_{h}^{Q},W_{h}^{K},W_{h}^{V}$ denote parameter matrices associated with the $h$ head. $H$ attention functions are applied in parallel to produce the output states $(o_{1}, o_{2}, ..., o_{H})$, then the $H$ output states are concatenated and linearly transformed to produce the final state. The process is formulated as:
\begin{equation}
o_{h} = Attention(Q_{h}, K_{h}, V_{h})
\end{equation}
\begin{equation}
O = [o_{1}, o_{2}, ..., o_{H}]W^{o}
\end{equation}
where $O$ is the final output states, and $W^{o}$ is a trainable matrix.

\section{Approach}
\label{sec:approach}
In this section, we first introduce the idea of stacking multiple shared layers, and then present the idea of closed parameters chain in detail.
\subsection{Stacking multiple shared layers}
\begin{figure*}[t]
	\centering
	\includegraphics[width=0.9\textwidth]{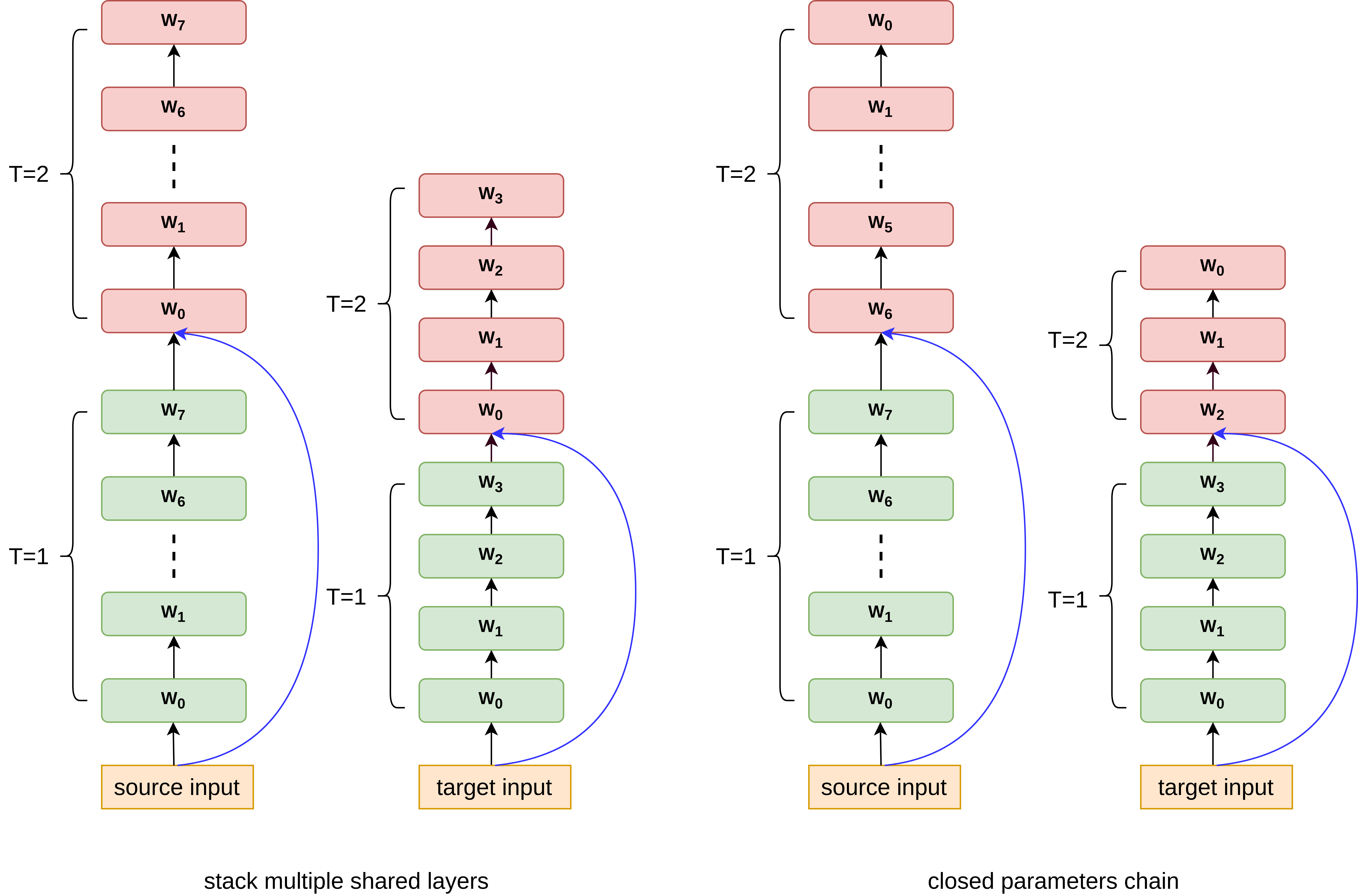} % Reduce the figure size so that it is slightly narrower than the column.
	\caption{The architecture of Transformer model variants. We assume that the size of recurrent $T$ is 2, the number of layers in the encoder is 8 and the number of layers in the decoder is 4. The rectangle denotes a layer in the encoder or decoder, and $w_{i}$ denotes the weight matrix of $i$-th layer. The output of the last layer in the encoder is also used as the input of each layer in the decoder, the output of the last layer in the decoder is used to generate the output sequence.}
	\label{fig1}
\end{figure*}
For vanilla Transformer, the encoder and decoder are both composed of a stack of 6 layers, and a stack of 6 layers is considered to be a block. each layer in the encoder consists of two sublayers and each layer in the decoder consists of three sublayers. We focus on the encoder to deepen the model, deepening the encoder requires fewer computational cost compared to the decoder. As figure \ref{fig1} shows, we choose to loop the encoder and decoder blocks to train a deeper model. At the same time, we reuse the source/target input as auxiliary information at the beginning of each block. The process is formulated as:
\begin{equation}
\begin{split}
&B_{t}^{Input} = B_{t-1}^{Output} + x \quad B_{t=0}^{Input} = x \\
&S_{i+1}^{t} = F(S_{i}^{t}) \\
&S_{0}^{t} = B_{t}^{Input} \quad B_{t}^{Output} = S_{N}^{t}
\end{split}
\end{equation}
where $x$ is the source/target input, $B_{t}^{Input}$ is the input of $t$-th block and $B_{t-1}^{Output}$ is the output of $(t-1)$-th block. $S_{i+1}^{t}$ is the hidden state of $(i+1)$-th position in $t$-th block, $N$ is the number of layers in each block. Function $F()$ consists of multi-head self-attention and feedforward neural network operations, pre-LN and dropout\cite{ott2019fairseq,srivastava2014dropout} are also used in each operation.

As the model deepens, the model parameters continue to increase. To reduce the model parameters, we share parameters in each block as well. In principle, our proposed approach can be applied to any multiple-layers neural models for training deeper model, such as Convs2s \cite{gehring2017convolutional} and Transformer \cite{vaswani2017attention}. In this work, we directly focus on Transformer considering that it achieves excellent performance on many translation tasks.

\subsection{closed parameters chain}
As figure \ref{fig1} shows, we design the parameters of each layer in the encoder and decoder to be a closed chain, we call it the closed parameters chain. Weight matrices of different layers can influence each other with the closed parameters chain strategy, we take the forward and backward process of the decoder as an example to present the idea of closed parameters chain, the detail is shown in figure \ref{fig2}. The forward process is formulated as follow:
\begin{figure*}[t]
	\centering
	\includegraphics[width=0.9\textwidth]{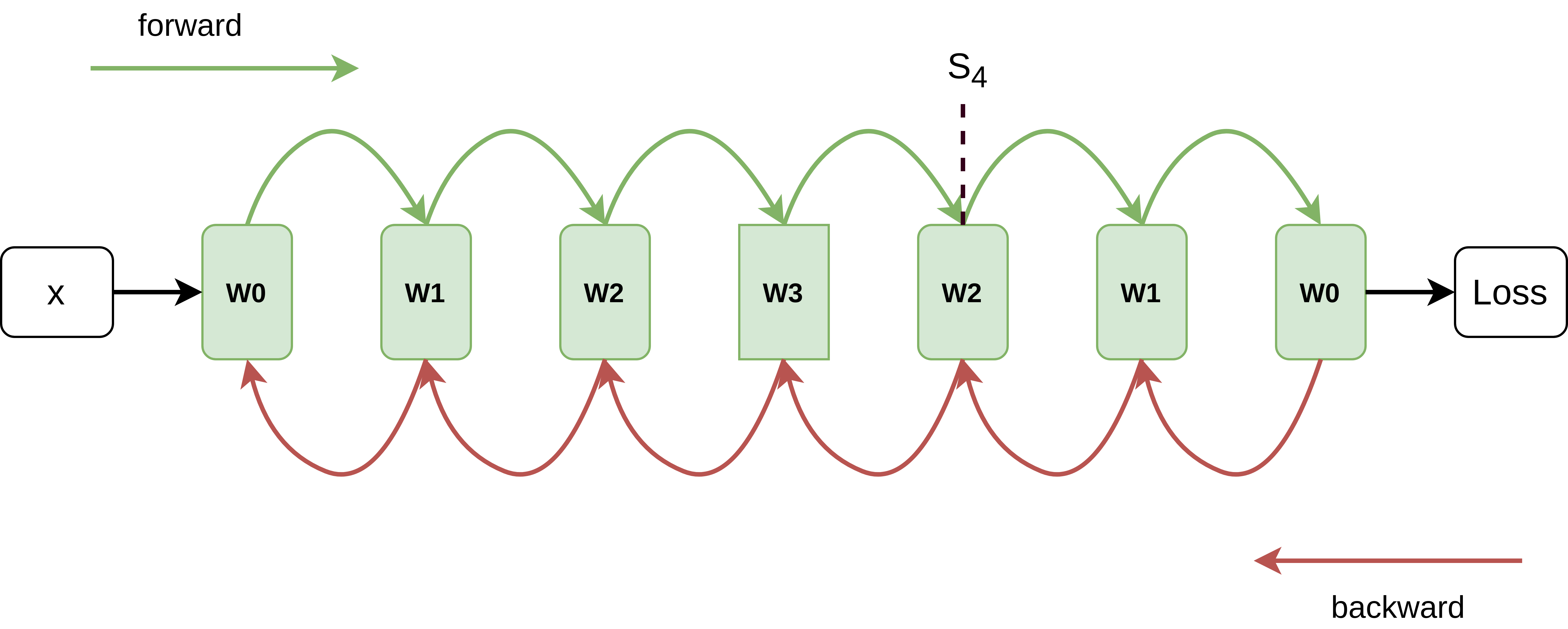} % Reduce the figure size so that it is slightly narrower than the column.
	\caption{The architecture of Tramsformer decoder variant. the green arrow denotes the forward process and the red arrow denotes the backward process.}
	\label{fig2}
\end{figure*}
\begin{equation}
\begin{split}
&B_{t=0}^{Input} = S_{0}^{t=0} = x \\
&B_{t=1}^{Input} = B_{t=0}^{Output} + x = S_{3}^{t=1} + x \\
&S_{n}^{t} = F(S_{n-1}^{t}) \\
&Loss = lossFunc(softmax(S_{6}^{t=1}), P_{gt})
\end{split}
\end{equation}
Here we assume the number of layers in the decoder to be 4$(N=4)$, the total layer of the decoder is 7, $S_{n}$ denotes the hidden state of $n$-th layer. we use $lossFunc()$ to calculate the final loss, and $P_{gt}$ is the real distribution. For the backward operation, we take the parameter $W_{1}$ and $W_{2}$ as an example, the update process is formulated as follow:
\begin{equation}
\begin{aligned}
&W_{1} = W_{1} - a * \frac{\partial{L}}{\partial{W_{1}}} \qquad W_{2} = W_{2} - a * \frac{\partial{L}}{\partial{W_{2}}} \\
\\
\frac{\partial{L}}{\partial{W_{1}}} &= \frac{\partial{Loss}}{\partial{S_{6}}} * \frac{\partial{S_{6}}}{\partial{S_{5}}} * \frac{\partial{S_{5}}}{\partial{W_{1}}} \\
& = \frac{\partial{Loss}}{\partial{S_{6}}} * W_{0} * (S_{4} + W_{1} * \frac{\partial{S_{4}}}{W_{1}}) \\
& = \frac{\partial{Loss}}{\partial{S_{6}}} * W_{0} * (S_{4} + W_{1} * W_{2}^{2} * W_{3} * S_{0}) \\
\frac{\partial{L}}{\partial{W_{2}}} &= \frac{\partial{Loss}}{\partial{S_{6}}} * \frac{\partial{S_{6}}}{\partial{S_{5}}} * \frac{\partial{S_{5}}}{\partial{S_{4}}} * \frac{\partial{S_{4}}}{\partial{W_{2}}} * \frac{\partial{S_{3}}}{\partial{S_{2}}} *
\frac{\partial{S_{2}}}{\partial{W_{2}}} \\
& = \frac{\partial{Loss}}{\partial{S_{6}}} * W_{0} * W_{1} * S_{3} * W_{3} * S_{1} \\
\end{aligned}
\end{equation}
Here, we set $F(x) = W*x$ and $W$ denotes the weight matrix. $Loss$ denotes the loss function and $a$ is the learning rate. Under the constraints of the closed parameter chain strategy, the gradient of $ W_ {1} $ and $ W_ {2} $ are interdependent.

\section{Experiment}

\subsection{Setup}
\subsubsection{Datasets and Evaluation}
Following the previous works, we evaluate our proposed model on WMT16 English-to-German(Dn-De) and WMT14 English-to-France(En-Fr) translation tasks.

For the En-De translation task, we use the same data set with \cite{ott2018scaling}'s work. The dataset contains about 4.5 million sentence pairs, with 116 million English words and 110 million German works. we validate on newstest2013 and test on newstest2014. The vocabulary is a 32k joint source and target byte pair encoding(BPE)\cite{sennrich2015neural}. For the En-Fr translation task, we use the significantly larger dataset consisting of 36 million sentence pairs, validate on newstest2012+2013 and test on newstest2014. The 40K vocabulary is also based on a joint source and target BPE factorization.

For evaluation,  we use beam search with a beam size of 4 and length penalty $\alpha=0.6$. At the same time, we also average checkpoints for all the models, and meansure the case-sensitive tokenized BLEU for all the translation tasks.
\subsubsection{Model and Hyperparameters}
All models were trained on the Fairseq \cite{ott2019fairseq} with 8 NVIDIA P100 GPUs, where each was allocated with a batch size of 8196 tokens. We employ the deep Transformer \cite{wang2019learning} as the baseline model, which consists of a 20-layers Encoder and 6-layers Decoder. For our base model, the hidden dimension $d_{x}=512$, the number of attention heads are 8, and the dimension of feed-forward inner-layer is 2048. For our proposed model, we keep the configuration the same with the baseline model, except that we adjust the number of layers in the Encoder and Decoder to 8 and 4. Unless otherwise noted, we set the number of loops in the Encoder and Decoder to be 5 and 2.

All models were optimized by Adam \cite{kingma2014adam} with $\beta_{1}=0.9, \beta_{2}=0.98$ and $\epsilon=10^{-9}$, the label smoothing $\epsilon_{ls}=0.1$ was used as regularization. We increase the learning rate for the first $warmup\_steps$ training steps and decrease it thereafter proportionally to the inverse square root of the step number, we used $warmup\_steps = 16000$. In generally, we followed the configure as suggested in Fairseq.
\subsection{Results}
\subsubsection{Results on the En-De task}
\begin{table*}[t]
\begin{center}
\begin{tabular}{ccccc}
	System & Architecture & Param & BLEU & $\Delta$ \\
	\hline
	\cite{wu2016google}& GNMT & - & 26.30 & -\\
	\cite{gehring2017convolutional} & ConvS2S & - & 26.36 & - \\
	\multirow{2}*{\cite{vaswani2017attention}} & Transformer-Base & 65M & 27.3 & - \\
	& Transformer-Big & 213M & 28.4 & - \\
	\multirow{2}*{\cite{he2018layer}} & Layer-wise Coordination-Base & - & 28.3 & +1.0 \\
	& Layer-wise Coordination-Big & $\dagger$210M & 29.0 & +0.6 \\
	\multirow{2}*{\cite{wang2019learning}} & Deep transformer(enc=20,dec=6) & 106M & 28.9 & +1.8$\ddagger$ \\
	& +DLCL(enc=30,dec=6) & 137M & 29.3 & +2.2$\ddagger$ \\
	\multirow{2}*{\cite{zhang2019improving}} & Transformer-Base + DS-Init & 72.3M & 27.50 & +0.3$\ddagger$\\
	& DS-Init + 20 layers + MAtt & 154.3M & 28.7 & +1.5$\ddagger$\\
	\hline
	\multirow{4}*{Our work} & Transformer-Base & 62M & 27.1 & refence \\
	& Transformer-Big & 211M & 28.45 & - \\
	& +Stack multiple shared layers (T=4,enc=8,dec=4) & 58M & 28.8 & +1.7 \\
	& +Stack multiple shared layers (T=4,enc=20,dec=4) & 96M & 29.3 & +2.2 \\
	& +Closed parameters chain (T=5,enc=8,dec=4) & 58M & 28.7 & +1.6 \\	
\end{tabular}
\end{center}
\caption{\label{table1} Comparing with existing NMT systems on WMT16 English-to-German translation task. "enc=8" denotes the layers of the  encoder is 8 and "dec=4" denotes the layers of the decoder is 4, "T" denotes the times of loop. $\dagger$ denotes an estimate value. $\ddagger$ denotes that this model has another baseline model. "Param" denotes the trainable parameter size of each model(M=million).}\smallskip
\end{table*}
In table \ref{table1}, we first show the results of the En-De translation task, which we compare to the existing systems based on self-attention. Obviously, most of the model variants outperform the Transformer significantly, although they have a larger amount of parameters. At the same time, the big-model usually has better model performance than the base-model, and it also requires more computational cost.

As for our approach, we first verify the idea of stacking multiple shared layers and closed parameters chain based on the Transformer-Base, where the encoder is composed of stack of 8 layers and the decoder is composed of stack of 4 layers. The results show that our proposed approaches outperform the Transformer-Base model significantly by over $1.7,1.6$ BLEU points,  and they have a similar size of model parameters compared with the Transformer-Base model. At the same time, our proposed approaches outperform the Transformer-Big model by over $0.35,0.25$ with a $27.49\%$ of Transformer-Big model parameters. Compared to the work of \cite{wang2019learning}, our proposed approaches based on the Transformer-Base model have $54.72\%$ model parameters and similar BLEU points.

To obtain a better model performance, we also apply the strategy of stacking multiple shared layers to the deep transformer, which has a 20-layer encoder and 6-layer decoder. Compared to \cite{wang2019learning}'s DLCL model, our model has equal BLEU points with only $70.1\%$ of model parameters.
\subsubsection{Results on the En-Fr task}
Seen from the En-De task, the RTAL based on the Transformer is more effective, where both the encoder and decoder have stack of 6 layers. Therefore, we evaluate our approach with the same structure on the En-Fr task. As seen in table \ref{table2}, we retrain the Transformer-Base and Transformer-Big models, and our approach outperforms the baseline models by 0.38 and 0.35 BLEU points. It confirms that our approach is a promising strategy for fusing information across layers in Transformer.
\begin{table}[t]
\begin{center}
\begin{tabular}{ccccc}
System & Architecture & Param & BLEU & $\Delta$ \\
\hline
\multirow{2}*{\cite{vaswani2017attention}} & Transformer-Base & 65M & 38.1 & refence \\
& Transformer-Big & 299M & 41.0 & - \\
\hline
\multirow{3}*{Our work} & Deep transformer (enc=20,dec=6) & 109M & 41.0 & +2.9 \\
& +Stack multiple shared layers (T=4,enc=8,dec=4) & 62M & 41.2 & +3.1 \\
& +Closed parameters chain (T=5,enc=8,dec=4) & 62M & 41.0 & +3.1

\end{tabular}
\caption{\label{table2} Experimental results on WMT17 English-to-France translation task. "Transformer-Base" denotes the Transformer-Base model. "Transformer-Big" denotes the Transformer-Big model.}\smallskip
\end{center}
\end{table}
\subsection{Analysis}
We conducted extensive analysis from different perspectives to better understand our model. All results are reported on the En-De translation task with Base-Model parameters, where the encoder and decoder are both composed of stack of 6 layers.
\subsubsection{Effect on Encoder and Decoder}
Both the encoder and decoder are multiple stacked layers, which may benefit from our proposed approach. In this experiment, we investigate the effects of RTAL on the encoder and decoder.
\begin{table}[h]
	\begin{center}
		\begin{tabular}{ccc}
			System & Aggregation position & BLEU \\
			\hline
			\multirow{3}*{T-Base + RTAL(6L)} & +encoder & 27.77 \\
			& +decoder & 27.82 \\
			& +both & 28.46
		\end{tabular}
		\caption{\label{table3} Experimental results of applying RTAL to different components on En-De test set. "+encoder" denotes that  RTAL is applied to the encoder. "+both" denotes that RTAL is applied to the encoder and decoder.}\smallskip
	\end{center}
\end{table}

As shown in table \ref{table3}, RTAL used in the encoder or decoder individually consistently outperforms the Transformer baseline model, and it further improves the model performance when using RTAL simultaneously in the encoder and decoder. These results claim that fusing information across layers is useful for both encoding the input sequence and generating output sequence.

\section{Conclusion}
In this paper, we propose a novel approach to fuse information across layers, which consists of a post-order binary tree and residual connection. To get better model performance, we also try many kinds of aggregation formulas. Our proposed approach is efficient and easy to implement. The experimental results show that our proposed approach outperforms the baseline model significantly.

\bibliographystyle{coling}
\bibliography{coling2020}

\end{document}